\documentclass[sigconf]{acmart}
\AtBeginDocument{%
  }

\setcopyright{acmcopyright}
\copyrightyear{2018}
\acmYear{2018}
\acmDOI{XXXXXXX.XXXXXXX}

\acmConference[Conference acronym 'XX]{Make sure to enter the correct
  conference title from your rights confirmation emai}{November 27--29,
  2018}{Woodstock, NY}
\acmPrice{15.00}
\acmISBN{978-1-4503-XXXX-X/18/06}

\begin{document}

\title{A Comparative Analysis of Fine-Tuned LLMs and Few-Shot Learning of LLMs for Financial Sentiment Analysis}


\author{Sorouralsadat Fatemi}
\email{sfatem6@uic.edu}
\affiliation{
  \institution{University of Illinois at Chicago}
  \country{USA}
}
\author{Yuheng Hu}
\email{yuhenghu@uic.edu}
\affiliation{%
  \institution{University of Illinois at Chicago}
  \country{USA}
}


\begin{abstract}
 Financial sentiment analysis plays a crucial role in uncovering latent patterns and detecting emerging trends, enabling individuals to make well-informed decisions that may yield substantial advantages within the constantly changing realm of finance.  Recently, Large Language Models (LLMs) have demonstrated their effectiveness in diverse domains, showcasing remarkable capabilities even in zero-shot and few-shot in-context learning for various Natural Language Processing (NLP) tasks. Nevertheless, their potential and applicability in the context of financial sentiment analysis have not been thoroughly explored yet. To bridge this gap, we employ two approaches: in-context learning (with a focus on gpt-3.5-turbo model) and fine-tuning LLMs on a finance-domain dataset. Given the computational costs associated with fine-tuning LLMs with large parameter sizes, our focus lies on smaller LLMs, spanning from 250M to 3B parameters for fine-tuning. We then compare the performances with state-of-the-art results to evaluate their effectiveness in the finance-domain. \\
Our results demonstrate that fine-tuned smaller LLMs can achieve comparable performance to state-of-the-art fine-tuned LLMs, even with models having fewer parameters and a smaller training dataset. Additionally, the zero-shot and one-shot performance of LLMs produces comparable results with fine-tuned smaller LLMs and state-of-the-art outcomes. Furthermore, our analysis demonstrates that there is no observed enhancement in performance for finance-domain sentiment analysis when the number of shots for in-context learning is increased.
\end{abstract}

\ccsdesc[500]{Computing methodologies~Natural language processing}

\keywords{Sentiment Analysis, Finance, Large Language Model, Natural Language Processing}


\maketitle

\section{Introduction}
The performance of various Large Language Models (LLMs) such as GPT3 \cite{1_brown2020language}, LLaMA \cite{2_touvron2023llama}, and T5 \cite{3_raffel2020exploring} has been remarkable across a diverse set of Natural Language Processing (NLP) tasks. Sentiment analysis is an NLP task that utilizes advancements in language modeling to achieve enhanced outcomes.  LLMs are trained on a wide range of corpora, enabling them to understand language patterns and perform tasks using zero-shot or few-shot prompting without explicit supervised training \cite{5_yang2023harnessing, 6_brown2020language}. Therefore, there exists considerable potential for the application of LLMs in sentiment analysis tasks, particularly in the financial domain where  a limited availability of annotated data.

Recent studies have highlighted the promising performance of LLMs in various NLP tasks, particularly in the zero-shot and few-shot settings. Notably, GPT-3 has shown competitive performance and, in some cases, even outperformed state-of-the-art models in few-shot in-context learning \cite{6_brown2020language,7_zhong2023can}. However, when it comes to sentiment analysis tasks, some studies have explored the zero-shot and in-context learning capabilities of LLMs and found that, in certain instances, in-context learning may not achieve comparable performance compared to fine-tuned and state-of-the-art models \cite{8_liu2022few, 34_wang2023chatgpt}. Therefore, it becomes crucial to investigate the zero-shot and few-shot in-context learning capabilities of LLMs in the financial domain, as it encompasses specific jargon and unique linguistic characteristics that necessitate special consideration. The aforementioned aspect has yet to be thoroughly examined in prior research, rendering it an important domain for further investigation.

Additionally, a number of studies indicate that in-context learning can yield inferior performance compared to fine-tuning \cite{6_brown2020language,8_liu2022few}. To delve further into this, we extend our experiments by fine-tuning LLMs on a finance-domain dataset and evaluate their performance on various financial datasets. For this purpose, we focus on a range of smaller language models known as FLAN-T5 models, with parameter sizes varying from 770 million to 3 billion parameters. Our goal is to determine whether their performance is comparable to state-of-the-art fine-tuned LLMs with larger parameter sizes. To comprehensively assess the performance of LLMs in financial sentiment analysis, we conduct comparisons between the fine-tuned models and their counterparts in zero-shot and few-shot settings.
In summary, our study involves conducting an empirical examination of the zero-shot and few-shot learning abilities of LLMs. In addition, we extend our investigation by performing fine-tuning on three Flan-T5 models (base, large, and xl) using finance-specific data. Additionally, we compare the results with state-of-the-art models to assess their performance comprehensively.

\section{related work} \label{section2}
Sentiment analysis in the financial domain has gained importance in NLP due to its impact on the stock prediction task. It plays a crucial role in providing valuable insights for investors' decision-making \cite{13_mishev2020evaluation}. previous literature explores diverse methodologies for conducting financial sentiment analysis, including lexicon-based techniques \cite{12_chen2018ntusd,14_loughran2011liability}. However, such approaches often struggle to capture the semantic nuances arising from specific word sequences. 

Alternatively, the advent of pre-trained language models, particularly models like BERT have revolutionized the field of NLP, including sentiment analysis \cite{6_brown2020language}. They leverage large-scale datasets and self-supervised learning to develop a robust understanding of context. This contextual awareness allows them to capture the subtle nuances and idiomatic expressions in sentiment-rich text \cite{16_devlin2018bert}. Although these models are trained on general domain corpora like news articles and Wikipedia, they lack specific knowledge in the finance domain. To address this limitation, researchers have undertaken further fine-tuning of BERT specifically using financial text, leading to promising outcomes in financial sentiment classification. Nevertheless, it is crucial to acknowledge that the fine-tuning procedure for language models such as BERT requires a substantial quantity of financial data and computational resources \cite{1_brown2020language,18_radford2019language}. In a specific study, BERT performed further pre-training by utilizing a financial communication corpus consisting of 4.9 billion finance-domain tokens \cite{18_radford2019language}.

This problem has been mitigated by the most recent advancements in NLP through the utilization of LLMs. These models are trained on vast amounts of textual data, spanning various domains and comprising hundreds of millions to billions of words, surpassing the scale of previous language models. In addition to leveraging larger amounts of pre-training data, large language models (LLMs) integrate various training techniques, including instruction tuning and reinforcement learning from human feedback (RLHF) \cite{21_wei2022chain,22_christiano2017deep}. This combination enables LLMs to achieve impressive performance in zero-shot or few-shot learning scenarios, occasionally surpassing the state-of-the-art benchmarks \cite{6_brown2020language}.

In light of these advancements, efforts have been made to develop financial-focused LLMs. One notable example is BloombergGPT, a 50 billion parameter language model specifically trained on a diverse range of financial data \cite{24_wu2023bloomberggpt}. Although, it demonstrated impressive performance in sentiment analysis, its closed-source nature limits its use in the research community. To address this, a new study proposed an open-source alternative called FinGPT, fine-tuned on extensive financial data from various sources \cite{25_yang2023fingpt}. FinGPT achieved comparable performance in financial sentiment analysis to BloombergGPT, with lower adaptation costs and computational requirements.

In a recent study, researchers introduced an instruction-tuned LLaMA-7B model, Instruct-FinGPT-7B, designed to overcome challenges faced by LLMs in accurately interpreting numerical values and comprehending financial context. The evaluation results showed that this model outperformed state-of-the-art methods in financial sentiment analysis, underscoring the promising potential of LLMs in this domain \cite{23_zhang2023instruct}. However, the focus of the present study is on fine-tuning smaller LLMs to achieve comparable performance while minimizing computational and memory resources.

Furthermore, there have been notable efforts to explore the zero-shot and in-context learning capabilities of LLMs in sentiment analysis tasks \cite{33_zhang2023sentiment, 34_wang2023chatgpt}. Results indicate that LLMs, like ChatGPT, exhibit robust zero-shot performance, even comparable to fine-tuned models like T5-large. However, they may not perform as well in few-shot learning, particularly with an increasing number of shots, especially on certain datasets like Twitter and MR \cite{33_zhang2023sentiment}. On the other hand, a separate study suggests that long-tail domains could benefit from few-shot prompting \cite{34_wang2023chatgpt}. In light of the varying outcomes observed in few-shot learning when applied to different sentiment analysis datasets, we aim to conduct a comprehensive examination to evaluate the effectiveness of zero-shot and few-shot learning methods for sentiment analysis in the finance-domain.

\section{Method and Experimental setup}
In this section, we outline the methodologies and datasets employed to investigate the objectives of our study.
\subsection{Zero-shot and Few-shot Settings}
In this investigation, we explore the zero-shot and few-shot learning capabilities of LLMs by conducting sentiment analysis inference without any specific training or modifying LLM parameters \cite{35_li2021prefix}. We use three models from Flan-T5 with different parameters, namely Flan-T5-Base (250M), Flan-T5-Large (780M), and Flan-T5-XL (3B), to study the impact of model size on zero-shot and few-shot learning performance. Additionally, we utilize the ChatGPT (gpt-3.5-turbo) model from OpenAI, known for its proficiency and cost-effectiveness within the GPT-3.5 family. To ensure consistency in our evaluations, we set the temperature parameter to zero, resulting in deterministic predictions. The performance of Flan-T5 models and ChatGPT is assessed under these conditions:
\begin{itemize}
\item {\textbf{zero-shot learning}}: The model is provided with only task name, task definition and label space, which serves as a set of options that enable the model to generate its response accordingly.
\item {\textbf{few-shot learning}}: The model is exposed to task definitions and input-output pairs containing K randomly selected labeled samples per class. We evaluated the few-shot learning capabilities of LLMs across three k-shot settings: 1-shot, 5-shot, and 10-shot. In each setting, we sampled K examples for each sentiment label.
\end{itemize}

\subsection{Fine-tuning LLMs} \label{section3.2}
For fine-tuning, we take three Flan-T5 models: Flan-T5 base (250M), Flan-T5-large (780M), and Flan-T5-xl (3B parameters). FLAN-T5 is an open-source finetuned version of Google's T5 model with instruct-finetuning \cite{11_chung2022scaling}.  They instruction-finetune T5 model which is a decoder-encoder model on a collection of data sources with a variety of instruction template types such as on different T5 model sizes.  The Flan-T5 series models have indeed demonstrated remarkable performance compared to T5 on specific benchmarks. For example, Flan-T5-XL, with only 3B parameters, achieved an impressive Massive Multi-task Language Understanding (MMLU) score of 52.4\%, surpassing the score of GPT-3 with 175B parameters by 8.5\% \cite{11_chung2022scaling}.

Furthermore, the authors evaluate Flan instruction tuning as an intermediate step before single target fine-tuning, aiming to determine whether Flan-T5 could serve as a superior starting checkpoint for subsequent fine-tuning. The results indicate that employing Flan-T5 as a starting checkpoint offers an additional advantage in terms of training efficiency. Additionally, during single target fine-tuning, Flan-T5 converges much faster than T5 and achieves higher accuracies (e.g., increased accuracy by 7.8\% in the RTE task). These findings strongly suggest that Flan-T5 is an excellent choice for further fine-tuning in our sentiment analysis task \cite{28_longpre2023flan}.

\subsection{Fine-tuning Procedure}
Fine-tuning pre-trained language models is a widely used technique to adapt LLMs for specific tasks by training them on task-specific data. However, this process becomes computationally intensive as it requires updating all pre-trained model parameters. \cite{29_hu2021lora}. To address this challenge, Microsoft researchers introduced a solution known as Low-Rank-Adaption (LoRA). LoRA introduces pairs of rank-decomposition weight matrices into the existing weights. During fine-tuning, only these newly added weights are trained, while the pre-trained model weights remain unchanged \cite{29_hu2021lora}. This method was further enhanced by the QLoRA technique, which involves backpropagating gradients through a frozen, 4-bit quantized pre-trained language model into LoRA \cite{30_dettmers2023qlora}. By integrating QLoRA, the fine-tuning process becomes more memory-efficient and computationally faster, while outperforming other efficient fine-tuning techniques \cite{30_dettmers2023qlora}.

Given that we are dealing with LLMs with 250M, 780M, and 3B parameters, we utilize the QLoRA method for fine-tuning. This approach, which offers enhanced memory efficiency and faster computation, is well-suited for optimizing the fine-tuning process for models of such sizes. Table 1 displays the reduced number of trainable parameters for each model. 

\section{performance evaluation}
In this section, we assess the efficacy of fine-tuning smaller LLMs and in-context learning of ChatGPT (GPT-3.5-turbo) and Flan-T5 models for financial sentiment analysis task. We compare the results with state-of-the-art approaches from previous studies, including FinBert and Instruct-FinGPT. In order to assess the effectiveness of each model, we report accuracy and F1-Macro for sentiment analysis tasks including fin-tuned models and in-context learning settings.

\subsection{Datasets}
The dataset used for fine-tuning Flan-T5 models is Twitter Financial News Sentiment (Twitter Train), which comprises tweets related to financial topics and is accessible through HuggingFace. The training dataset consists of 9540 samples, with each sample labeled as Positive, Negative, or Neutral based on the sentiment expressed in the tweets. 

Testing dataset consists of two datasets (all datasets are available through Hugging Face):

\begin{itemize}
\item {\textbf{Twitter financial news sentiment validation (TFSN)}}: The dataset comprises 2,390 samples of an annotated corpus of finance-related tweets. Each sample is labeled as Positive, Negative, or Neutral.
\item {\textbf{Financial PhraseBank (FPB)}}: The dataset consists of samples randomly extracted from financial news articles annotated by a team of 16 domain experts. Additionally, the dataset includes information on the agreement levels among the annotators for each sentence. We select the sentences with a 50\% agreement level between annotators which consists of 4845 financial news and their sentiment label of positive, negative, or
neutral \cite{32_malo2014good}.
\end{itemize}
 We utilize the entire training dataset for fine-tuning. Subsequently, we perform inference for fine-tuned model, and in-context settings on both test datasets.
 
\subsection{Model Training Details}

For each Flan-T5 model (base, large, and xl), we follow the same fine-tuning procedure with the same hyperparameters. To compress the pretrained language models, we load the Flan-T5 models in a 4-bit format. To further reduce memory requirements during fine-tuning, we employ LoRa with an attention dimension (r) of 8, an alpha parameter of 32, and a dropout probability of 0.05, resulting in a reduction in the number of trainable parameters, as illustrated in Table \ref{table:1}.
\begin{table}[h!]
\begin{tabular}{ c c c } 
\hline
Model & All Parameters & Trainable Parameters \\
\hline
Flan-T5-Base & 248M & 0.88M \\
Flan-T5-Large & 785M & 2.36M \\
Flan-T5-XL & 2.85B & 4.72M \\
\hline
\end{tabular}
\caption{Number of trainable parameters after using QLoRA.}
\label{table:1}
\end{table}
Then, we perform fine-tuning over 3 epochs with a learning rate of 1e-4 and a maximum input text length of 256 tokens. To prevent CUDA Out-of-Memory errors, we set the batch size to fit into memory automatically through exponential decay.  We conduct fine-tuning using one A100 GPU, which leads to the following total training times for each model: 28 minutes for the base model, 54 minutes for the large model, and 65 minutes for the XL model. 

\begin{figure*}
    \centering
    \includegraphics[scale=0.8]{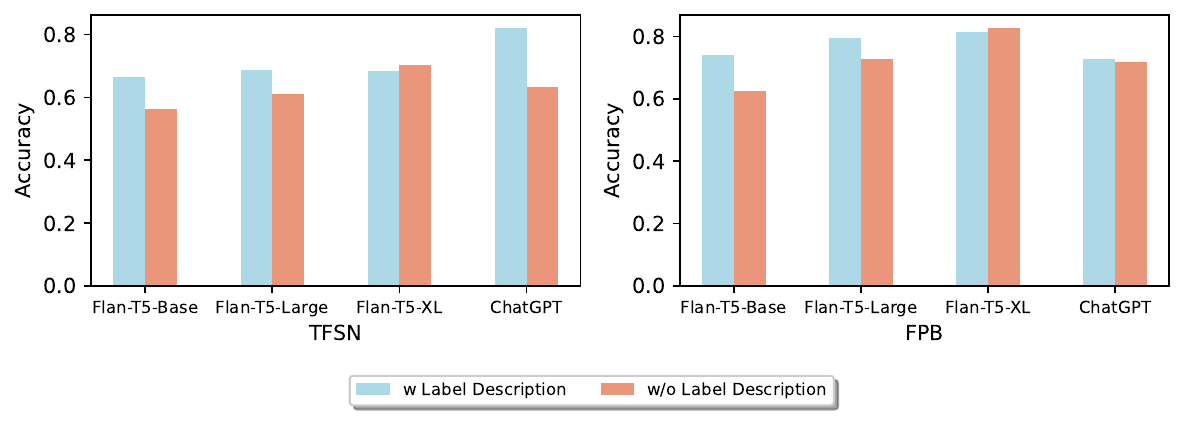}
    \caption{Zero-shot performance of Flan-T5 models and ChatGPT with and without label description in the prompt.}
    \label{fig:1}
\end{figure*}

\begin{table*}
\begin{tabular}{ c c | c c | c | c } 
\hline
Dataset & Metric 
& 
FinBert  & Instruct-Llama-7B
& 
\begin{tabular}{ c }
Fine-tuned-Flan-T5   \\ 
\begin{tabular}{ c c c  }
Base & Large & XL 
\end{tabular}
\end{tabular}
& 
\begin{tabular}{ c }
ChatGPT  \\ 
\begin{tabular}{ c c c c }
0-shot & 1-shot & 5-shot & 10-shot
\end{tabular}
\end{tabular}
 \\
\hline
FPB &
\begin{tabular}{ c }
Acc. \\ F1
\end{tabular}
& 
\begin{tabular}{ c }
-- \\ --
\end{tabular}
& 
\begin{tabular}{ c }
0.758 \\ 0.739
\end{tabular}
& 
\begin{tabular}{ c c c }
0.769 & 0.793 & \textbf{0.807} \\ 0.726 & 0.759 & 0.780
\end{tabular}
&
\begin{tabular}{ c c c c }
0.727  &  0.791 &  0.795 & 0.779  \\  0.680   &  0.774  &  \textbf{0.794}  & 0.783 
\end{tabular}
\\ \hline
TFSN &
\begin{tabular}{ c }
Acc. \\ F1
\end{tabular}
& 
\begin{tabular}{ c }
0.725 \\ 0.668
\end{tabular}
& 
\begin{tabular}{ c }
0.881 \\ 0.842
\end{tabular}
& 
\begin{tabular}{ c c c }
0.874 & 0.894 & \textbf{0.903} \\ 0.838 & 0.867 & \textbf{0.878}
\end{tabular}
&
\begin{tabular}{ c c c c }
0.821  & 0.823 &  0.768 &  0.724 \\ 0.818   &  0.819 & 0.775 & 0.732
\end{tabular}
\\ \hline
\end{tabular}

    \caption{Performance comparison between fine-tuned Flan-T5 models, in-context learning of ChatGPT, and state-of-the-art models. Instruct-FinGPT-7B \cite{23_zhang2023instruct} results are taken from its respective papers. FinBERT results for the FPB dataset are not reported due to its use as the training set. The best results are in \textbf{bold}.}
\label{table:2}
\end{table*}

\subsection{Benchmark Models}
In this section, we present outcomes obtained from previous studies that have demonstrated superior results.
\begin{itemize}

\item{\textbf{FinBERT:}} This model is a variant of BERT, pre-trained on a financial text corpus comprising 1.8M news articles from Reuters TRC2 dataset. For inference, the model was accessed via the HuggingFace API pipeline. The sentences are initially tokenized and then fed into FinBERT for inference. FinBERT generates sentiment analysis outcomes for each textual input, classifying them as positive, negative, or neutral.

\item{\textbf{Instruct-FinGPT-7B:}} The model is fine-tuned on LLaMA-7B model, employing the instruction tuning dataset. For evaluation purposes, we rely on the results reported in the paper that originally introduced Instruct-FinGPT-7B \cite{23_zhang2023instruct}.
\end{itemize}

\subsection{Prompt Design}
The effectiveness and coherence of prompt formats can vary based on the training methodology and data used during the training process. In our experiments, we started with a straightforward prompt for zero-shot inference, as shown below:
\begin{itemize}
\item {\texttt{You are an AI language model trained to detect the sentiment of each sentence for stock prediction. Analyze the following sentence and determine if the sentiment is: positive or negative or neutral. Return only a single word, either Positive or Negative or Neutral.}}
\end{itemize}

Subsequently, we introduced label descriptions along with various prompt formats\footnote{Examples of prompt formats can be found at the following link:\url {https://help.openai.com/en/articles/6654000-best-practices-for-prompt-engineering-with-openai-api}}, such as """ to separate the instruction and context, to assess whether they could enhance the model's performance. The example of zero-shot and one-shot prompt is shown in Figure \ref{fig:1}.

We evaluated the performance of all three Flan-T5 models and the ChatGPT model using zero-shot inference. The results, shown in Figure 1, indicate higher accuracy across most models with the prompt containing label descriptions compared to the prompt without label descriptions. Therefore, we adopted the prompt format shown in Table \ref{table: 5} for all zero-shot and few-shot settings inference.

\begin{table}
\begin{center}
\begin{tabular}{ | c | c | c | } 
\hline
Model & 
\begin{tabular}{ c } 
TFSN \\
\hline
\begin{tabular}{ c c } 
Acc. & F1
\end{tabular}
\end{tabular}
&
\begin{tabular}{ c } 
FPB \\
\hline
\begin{tabular}{ c c } 
Acc. & F1
\end{tabular}
\end{tabular}
\\
\hline 
\begin{tabular}{ c } 
Base \\ \hline 
0-Shot \\ 1-Shot \\ 5-Shot \\ 10-Shot
\end{tabular}
& 
\begin{tabular}{ c c } 
\begin{tabular}{ c } 
\\ 0.665 \\ 0.694 \\ 0.707 \\0.726
\end{tabular}
& 
\begin{tabular}{ c } 
\\ 0.593 \\ 0.624 \\ 0.643 \\ 0.659
\end{tabular}
\end{tabular}
& 
\begin{tabular}{ c c } 
\begin{tabular}{ c } 
\\ 0.741 \\ 0.726 \\ 0.734 \\ 0.764
\end{tabular}
& 
\begin{tabular}{ c } 
\\ 0.729 \\ 0.727 \\ 0.742 \\ 0.742
\end{tabular}
\end{tabular}
\\ \hline
\begin{tabular}{ c } 
Large \\ \hline
0-Shot \\ 1-Shot \\ 5-Shot \\ 10-Shot
\end{tabular}
& 
\begin{tabular}{ c c } 
\begin{tabular}{ c } 
\\ 0.687 \\ 0.742 \\ 0.765 \\0.778
\end{tabular}
& 
\begin{tabular}{ c } 
\\ 0.647 \\ 0.688 \\ 0.701 \\ 0.704
\end{tabular}
\end{tabular}
& 
\begin{tabular}{ c c } 
\begin{tabular}{ c } 
\\ 0.797 \\ 0.809 \\ 0.813 \\ 0.804
\end{tabular}
& 
\begin{tabular}{ c } 
\\ 0.798 \\ 0.807 \\ 0.810 \\ 0.801
\end{tabular}
\end{tabular}
\\ \hline
\begin{tabular}{ c } 
XL \\ \hline
0-Shot \\ 1-Shot \\ 5-Shot \\ 10-Shot
\end{tabular}
& 
\begin{tabular}{ c c } 
\begin{tabular}{ c } 
\\ 0.683 \\ 0.791 \\ 0.760 \\0.771
\end{tabular}
& 
\begin{tabular}{ c } 
\\ 0.678 \\ 0.753 \\ 0.730 \\ 0.729
\end{tabular}
\end{tabular}
& 
\begin{tabular}{ c c } 
\begin{tabular}{ c } 
\\ 0.815 \\ 0.847 \\ 0.832 \\ 0.780
\end{tabular}
& 
\begin{tabular}{ c } 
\\ 0.823 \\ 0.836 \\ 0.833 \\ 0.796
\end{tabular}
\end{tabular}
\\ \hline
\end{tabular}
\end{center}
    \caption{Few-shot performance of Flan-T5 models on TFSB and FPB datasets.}
\label{table:3}
\end{table}

\begin{figure*}
    \centering
    \includegraphics[scale=0.7]{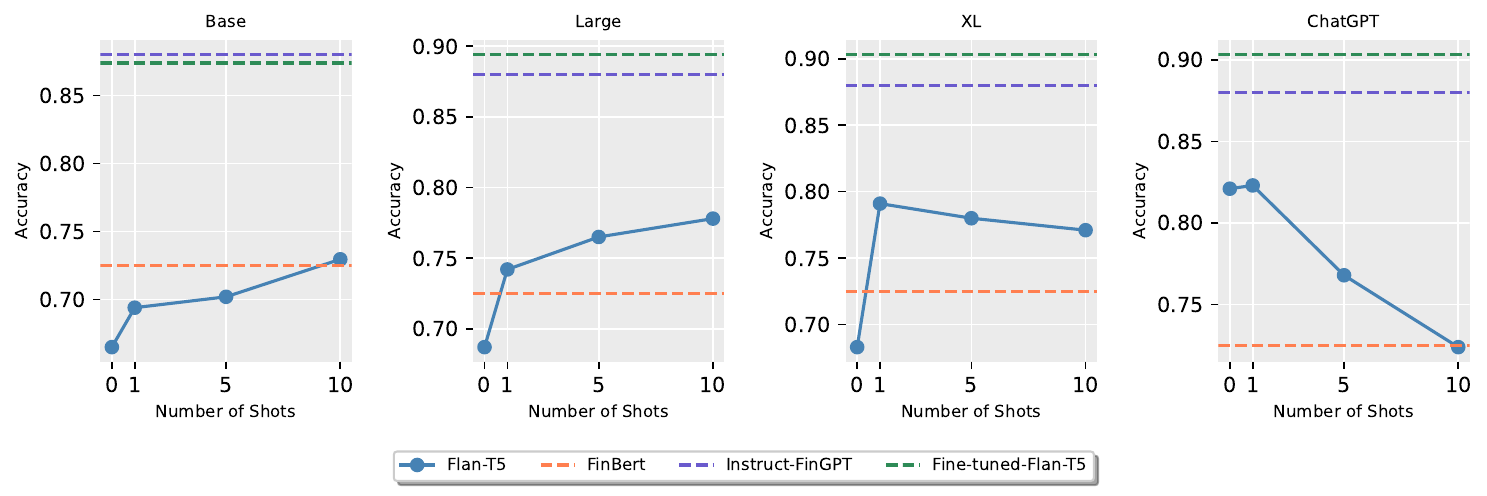}
    \caption{Few-shot prompting results on TFSN dataset compared with FinBERT, Fine-tuned-Flan-T5, and Instruct-FinGPT results. Utilizing the best-performing fine-tuned Flan-T5-XL model for ChatGPT.}
    \label{fig:2}
\end{figure*}

\begin{figure*}
    \centering
    \includegraphics[scale=0.7]{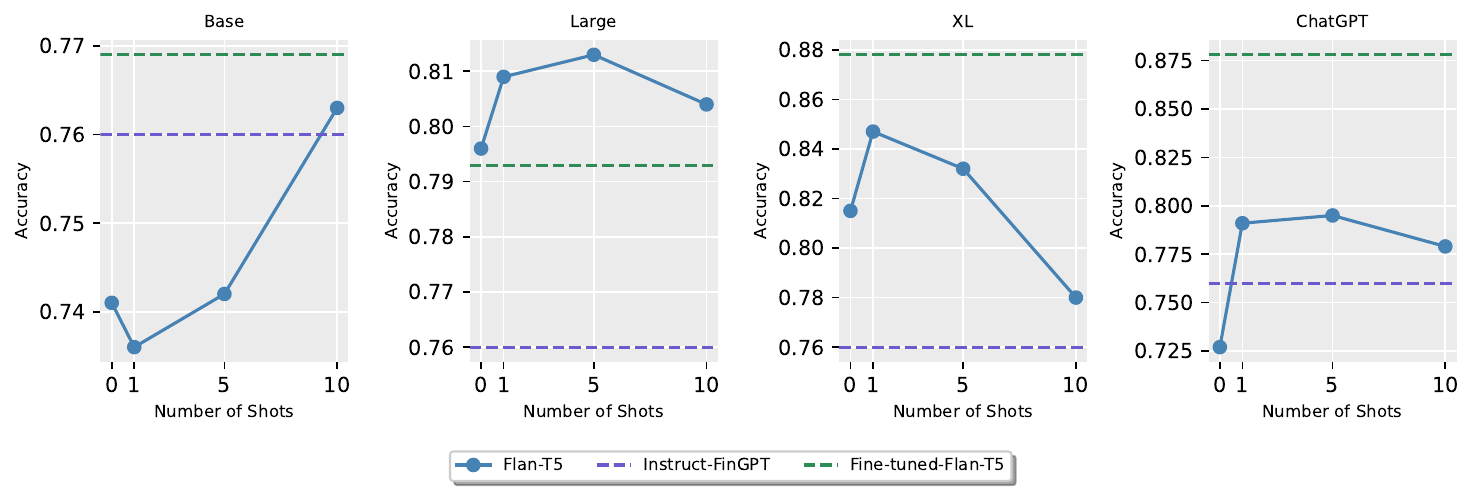}
    \caption{Few-shot prompting results on FPB dataset compared with Fine-tuned-Flan-T5 and Instruct-FinGPT results. Utilizing the best-performing fine-tuned Flan-T5-XL model for ChatGPT.}
    \label{fig:3}
\end{figure*}

\section{Evaluation Results}
As mentioned in section \ref{section2}, our main focus is on evaluating the zero-shot and few-shot learning performance of LLMs, including both smaller and larger models in terms of parameter size, in comparison to fine-tuned smaller LLMs. The results of Flan-T5 fine-tuned models on both TFSN and FPB datasets, along with the benchmark models, and the zero-shot and few-shot performance of ChatGPT, are presented in Table \ref{table:2}. Additionally, the results of the zero-shot and few-shot performance of Flan-T5 models are shown in Table \ref{table:3}.

\textbf{Fine-tuned LLMs Results.} As the results indicate in Figure \ref{fig:2} and Table \ref{table:2}, on the TFSN dataset, the performance of fine-tuned Flan-T5 models is comparable to the state-of-the-art model, Instruct-FinGPT ,and significatly outperforms FinBert results. It is noteworthy that we achieved this level of performance with significantly fewer computational resources (1 A100 GPU compared to 8 A100 GPUs) and a comparable or even shorter training time by utilizing QLoRA method , especially for the Flan-T5-Base model (28 minutes). These findings align with a previous study that highlights the efficiency advantage of using Flan-T5 as a starting checkpoint for further fine-tuning, as discussed in section \ref{section3.2}.

\textbf{Zero-shot Results.} As shown in Figure \ref{fig:2}, the zero-shot learning results of Flan-T5 models (Base, Largr, and XL) on the TFSN dataset fall significantly behind those of all fine-tuned models (FinBert, Instruct-FinGPT, and all fine-tuned Flan-T5 models). Notably, Instruct-FinGPT and fine-tuned Flan-T5 models outperform LLMs by a clear margin of roughly 20\%. However, the zero-shot performance of ChatGPT reaches 82\%, surpassing the FinBert model but still remaining inferior to Instruct-FinGPT and all fine-tuned Flan-T5 models.

As depicted in Figure \ref{fig:3} and Table \ref{table:3}, the zero-shot results of all Flan-T5 models on the FPB dataset show more promising outcomes, even performing comparably to the corresponding fine-tuned Flan-T5 models. This observation aligns with previous research findings \cite{33_zhang2023sentiment}. Additionally, a noteworthy finding on the FPB dataset is that larger models, with a greater number of parameters, tend to outperform the smaller ones in zero-shot inference. For instance, when comparing the performance between Flan-T5-Base and Flan-T5-XL, there is an increase of roughly 7\% in performance.

When comparing the zero-shot performance of Flan-T5 models to ChatGPT, ChatGPT appears to be less accurate despite having larger parameters, which contradicts the zero-shot inference on the TFSN dataset. This discrepancy in performance can be explained by the different language and jargon used in the FPB news headlines and the TFSN social media posts. ChatGPT, being trained on a large corpus of social media posts, is better equipped to extract the true sentiment from social media text, which is reflected in its zero-shot performance on the TFSN dataset. On the other hand, smaller LLMs like Flan-T5 models outperform ChatGPT in the zero-shot setting on the FPB dataset, possibly due to their large-scale instruction tuning, which enhances their reasoning capabilities and allows them to extract sentiment from more complex texts.

\textbf{Few-shot Results.} Results comparing few-shot prompting performance with zero-shot prompting performance are presented in Figure \ref{fig:2}, Figure \ref{fig:3}, and Table \ref{table:3}. Notably, we observe that 1-shot prompting can significantly improve the performance across almost all models and datasets, except for Flan-T5-Base (250M), where there is a slight decrease in performance. This decrease can be attributed to the impact of a single unrelated example, which can have an adverse effect on the smaller LLMs. Interestingly, we find that smaller LLMs (Flan-T5) benefit more on average from one demonstration compared to ChatGPT.

The impact of increasing shots to 5 and 10 exhibited variability across various models and datasets. On the TFSN dataset, it was observed that a marginal improvement in performance was achieved for both Flan-T5-Base and Flan-T5-Large models with an increase in the number of shots. However, the performance of Flan-T5-Large and Flan-T5-XL was impeded by the increase in shots. The variation in outcomes can be ascribed to the difficulties associated with managing excessively lengthy contexts, which have the potential to lead the LLMs misguided, as revealed in a new study \cite{33_zhang2023sentiment}.

These findings are consistent with a previous study that highlighted the sensitivity issue of LLMs when exposed to few-shot examples during inference \cite{33_zhang2023sentiment}. In order to mitigate these disparities and obtain consistent enhancements in performance, it is imperative to employ more efficacious techniques for few-shot learning. For example, a recent study suggested the retrieval of examples that possess semantic similarity to a test query sample in order to generate its corresponding prompt \cite{36_liu2021makes}. Moreover, the introduction of the Chain of Thought (CoT) and Clue And Reasoning Prompting (CARP) methods aimed to improve the reasoning capabilities of LLMs. which could be beneficial for extracting sentiment from financial corpora \cite{37_sun2023text, 38_wei2022chain}. Our future work will encompass an exploration of these approaches

\section{Conclusion and future directions}

In this study, we extensively compared the performance of fine-tuned LLMs with different parameter sizes (ranging from 250M to 3B) and their in-context learning capabilities, for financial sentiment analysis. Our experimental results demonstrate that fine-tuned LLMs achieved comparable performance to state-of-the-art models while utilizing significantly fewer computational resources. The zero-shot and one-shot settings performed impressively, especially on the FPB dataset, which can be attributed to the corpus used for pre-training the LLMs. Moreover, larger LLMs demonstrated better performance in the zero-shot and one-shot settings.

The remarkable results obtained from fine-tuned, zero-shot, and one-shot inferences of Flan-T5 models can be attributed to their instruct fine-tuning. However, we observed inconsistent performance in the five-shot and ten-shot settings across all models and datasets. Notably, increasing the number of shots did not lead to improved performance on average. Nevertheless, the models demonstrated reasonable performance on both datasets and across all three Flan-T5 models and ChatGPT, indicating their potential for financial sentiment analysis considering the scarcity of labeled data in finance-domain.

In conclusion, our study showcases the remarkable capabilities of LLMs, even smaller models, in both fine-tuning and in-context learning for financial sentiment analysis task. These findings provide valuable insights into potential avenues for further investigation in the field of financial sentiment analysis. 

For future studies, we plan to assess the new methods of prompt formatting, such as CoT and CARP, and prompt selection by retrieving semantically-similar examples to the test queries, rather than random sampling for few-shot settings \cite{37_sun2023text, 38_wei2022chain}. The objective of this study is to determine the extent to which these methods can consistently enhance the performance of LLMs in financial sentiment analysis tasks.

\bibliographystyle{ACM-Reference-Format}
\bibliography{sample-base}

\appendix
\section{Appendices}
We introduce a zero-shot and 1-shot prompt designed for the financial sentiment analysis task, displayed on the subsequent page.
\begin{table*}
\begin{tabular}{|l|}
\hline
Zero-shot prompt \\
\hline \\
 
 {\small You are an AI language model trained to detect the sentiment of sentences for stock prediction
See below all the possible labels and their descriptions} \\
"""\\
\small description: Bearish sentiment\\
\small label: Negative\\
"""\\
\small description: neutral sentiment\\
\small label: Neutral\\
"""\\
\small description: Bullish sentiment\\
\small label: Positive\\
"""\\
\small Here is the sentence that needs to be classified\\
\small sentence: {\fontfamily{qcr}\selectfont\{sentence\}}\\
\small label:\\
\hline
\hline
One-shot Prompt \\
  \hline 
   {\small You are an AI language model trained to detect the sentiment of sentences for stock prediction
See below all the possible labels and their descriptions} \\
\small """\\
\small description: Bearish sentiment\\
\small label: Negative\\
\small """\\
\small description: neutral sentiment\\
\small label: Neutral\\
\small """\\
\small description: Bullish sentiment\\
\small label: Positive\\
\small """\\
\small See below a couple of examples\\
\small """\\
\small text: Cemex cut at Credit Suisse, J.P. Morgan on weak building outlook \\
\small label: Negative\\
\small """\\
\small text: ululemon Falls on Conservative View But Analysts Keep Faith\\
\small label: Neutral\\
\small """\\
\small text: Wells Fargo Downgrades Netflix \$NFLX to Underperform but sees as a takeover target. NFLX could get acquired\\
\small label: Positive\\
\small"""\\

\small Here is the sentence that needs to be classified\\
\small sentence: {\fontfamily{qcr}\selectfont\{sentence\}}\\
\small label:\\
\hline
\end{tabular}
\caption{Prompts for zero-shot and one-shot settings.  In the Five-Shot and Ten-Shot settings, 5 and 10 samples were added for each class, respectively.}
\label{table: 5}
\end{table*}






\end{document}